%% file: IEEE_main.tex
\documentclass[sigconf]{acmart}

\input{macro}
\begin{document}
\title{\textit{ClassEval}: A Manually-Crafted Benchmark \\ for Evaluating LLMs on Class-level Code Generation}

\renewcommand{\shorttitle}{\textit{ClassEval}: A Manually-Crafted Benchmark for Evaluating LLMs on Class-level Code Generation}
\author{Xueying Du \quad Mingwei Liu \quad Kaixin Wang \quad Hanlin Wang \quad Junwei Liu \\ Yixuan Chen \quad Jiayi Feng \quad Chaofeng Sha \quad Xin Peng \quad Yiling Lou }

\affiliation{%
  \institution{Fudan University}
  \country{ Shanghai, China}
}
\email{{xueyingdu21, kxwang23, wanghanlin23}@m.fudan.edu.cn}
\email{{jwliu22, 23212010005, 23210240148}@m.fudan.edu.cn}
\email{{liumingwei, cfsha, pengxin, yilinglou}@fudan.edu.cn}

\settopmatter{printacmref=false}

\renewcommand\footnotetextcopyrightpermission[1]{} 
%\author{Anonymous}
\begin{abstract}
\input{sections/abstract}
\end{abstract}

\maketitle

\input{sections/introduction}
\input{sections/background}

\input{sections/approach}

\input{sections/evaluation/evaluation}

\input{sections/threat}

\input{sections/related}
\input{sections/conclusion}

\balance
\bibliographystyle{IEEEtran}
\bibliography{ref}

\end{document}

%% file: macro.tex
\usepackage{adjustbox}
\usepackage{amsmath,amssymb,amsfonts}
\usepackage[linesnumbered,ruled,vlined]{algorithm2e}
\usepackage{setspace}
\usepackage{algorithmic}
\usepackage{float}
\usepackage{graphicx}
\usepackage{textcomp}
\usepackage{xcolor}
\usepackage{multirow}
\usepackage{mdframed}
\usepackage{balance}
\usepackage{booktabs}
\usepackage{tcolorbox}
\usepackage{makecell}
\usepackage{threeparttable}
\usepackage{colortbl}
\usepackage{subfigure}
\usepackage{hhline}
\usepackage{pifont}
\usepackage{listings}
\usepackage{enumitem}
\newcommand{\tabincell}[2]{\begin{tabular}{@{}#1@{}}#2\end{tabular}}

\newcommand{\parabf}[1]{\noindent\textbf{#1}}

\definecolor{ggray}{HTML}{eff0f0}
\definecolor{gggray}{HTML}{E8E8E8}
\definecolor{ggggray}{HTML}{BEBEBE}

\newcommand{\ourbenchmark}{ClassEval\xspace}

\newcommand{\ie}{\textit{i.e.,}\xspace}
\newcommand{\eg}{\textit{e.g.,}\xspace}
\newcommand{\passk}{\textit{Pass@k}\xspace}
\newcommand{\passfive}{\textit{Pass@5}\xspace}
\newcommand{\passone}{\textit{Pass@1}\xspace}

\newcommand{\dep}{\textit{DEP}\xspace}

\newcommand{\gptthree}{GPT-3.5\xspace}
\newcommand{\gptfour}{GPT-4\xspace}

\newcommand{\inscodegen}{Instruct-CodeGen\xspace}

\newcommand{\wizardcoder}{WizardCoder\xspace}
\newcommand{\starcoder}{Instruct-StarCoder\xspace}
\newcommand{\codegeex}{CodeGeeX\xspace}
\newcommand{\incoder}{InCoder\xspace}
\newcommand{\polycoder}{PolyCoder\xspace}

\newcommand{\mbpp}{MBPP\xspace}
\newcommand{\chatglm}{ChatGLM\xspace}
\newcommand{\santacoder}{SantaCoder\xspace}

\newcommand{\humaneval}{HumanEval\xspace}

\newcommand{\topicmath}{Mathematical Operations\xspace}
\newcommand{\topicnlp}{Natural Language Processing\xspace}
\newcommand{\topicdataprocess}{Data Formatting\xspace}
\newcommand{\topicfile}{File Handling\xspace}
\newcommand{\topicdatabase}{Database Operations\xspace}
\newcommand{\topicproject}{Management Systems\xspace}
\newcommand{\topicgame}{Game Development\xspace}
\newcommand{\holisticgen}{Holistic\xspace}
\newcommand{\incrementalgen}{Incremental\xspace}
\newcommand{\individualgen}{Compositional\xspace}

\definecolor{myyellow}{HTML}{FFF2CC}

\usepackage[utf8]{inputenc}
\usepackage[english]{babel}

\newcounter{finding}
\newcommand{\finding}[1]{\refstepcounter{finding}
 	\vspace{1mm}
	\begin{mdframed}[linecolor=gray!25,roundcorner=12pt,backgroundcolor=myyellow!30,linewidth=3pt,innerleftmargin=2pt, leftmargin=0cm,rightmargin=0cm,topline=false,bottomline=false,rightline = false]
		\textbf{Finding \arabic{finding}:} #1
	\end{mdframed}
	\vspace{1mm}
}

\usepackage{footmisc}

\newcommand{\distance}{5pt}

%% file: sections/abstract.tex
Recently, many large language models (LLMs) have been proposed, showing advanced proficiency in code generation. Meanwhile, many efforts have been dedicated to evaluating LLMs on code generation benchmarks such as HumanEval. Although being very helpful for comparing  different LLMs, existing evaluation  focuses on a \textit{simple} code generation scenario (\ie{} function-level or statement-level code generation), which mainly asks LLMs to generate one single code unit (\eg{} a function or a statement) for the given natural language description. Such evaluation focuses on generating independent and often small-scale code units, thus leaving it unclear how LLMs perform on generating more complicated code. 

To fill this knowledge gap, we make the first attempt to evaluate LLMs in a more \textit{challenging} code generation scenario, \ie{} class-level code generation. We first manually construct the first class-level code generation benchmark \ourbenchmark{} of 100 class-level Python code generation tasks with approximately 500 person-hours. Based on the new benchmark \ourbenchmark{}, we then perform the first study of 11 state-of-the-art LLMs on class-level code generation. Based on our results, we have the following main findings. First, we find that all existing LLMs show much worse performance on class-level code generation compared to on standalone method-level code generation benchmarks like \humaneval{}; and the method-level coding ability cannot equivalently reflect the class-level coding ability among LLMs. Second, we find that \gptfour{} and \gptthree{} still exhibit dominate superior than other LLMs on class-level code generation, and the second-tier models includes \starcoder{}, \inscodegen{}, and \wizardcoder{} with very similar performance. Third, we find that generating the entire class all at once (\ie{} holistic generation strategy) is the best generation strategy only for \gptfour{} and \gptthree{}, while method-by-method generation (\ie{} incremental and compositional) is better strategies for the other models with limited ability of understanding long instructions and utilizing the middle information. Lastly, we find the limited model ability of generating method-dependent code and discuss the frequent error types in generated classes. Our benchmark is available at https://github.com/FudanSELab/ClassEval

%% file: sections/introduction.tex
\section{Introduction}
Code generation techniques automatically generate code snippets for the given natural language description, which can be leveraged to improve development productivity and have been extensively studied in literature~\cite{vikram2023large,10172763,kang2023explainable}. The recent advance in large language models (LLMs) has brought significant advancements in the code generation domain. 
To date, researchers have proposed various LLMs~\cite{openai2023gpt4, bian2023chatgpt, zheng2023vicuna, du2022chatglm, luo2023wizardcoder, li2023starcoder, intructcodegen, zheng2023codegeex, xu2022polycoder, fried2022incoder, allal2023santacoder}
% \yiling{add citation of all studied models}
(such as GPT-4~\cite{openai2023gpt4}, \wizardcoder{}~\cite{luo2023wizardcoder}, and \inscodegen~\cite{intructcodegen}) by training large models with over billions of parameters on massive general or code-specific corpora and instructions. 
 
To fully understand the code generation capability of emerging LLMs, many efforts have been dedicated to evaluating LLMs on automatically or manually constructed code generation benchmarks. To date, many code generation benchmarks have been proposed, such as \humaneval{}~\cite{chen2021huamneval} and \mbpp{}~\cite{austin2021mbpp}. Although being very helpful for people to understand and compare the performance of different LLMs, existing evaluation actually focuses on a rather \textit{simple} code generation scenario, \ie{} function-level or statement-level code generation. They mainly ask LLMs to generate one single code unit (\eg{} a function or a statement) for the given natural language descriptions in a standalone way, which inherently have two limitations in evaluating LLMs in code generation. 
\textit{First, such evaluation tends to focus on generating code of short length}, \eg{}  each task in the most widely-used benchmark \humaneval{} only involves generating code of 11.5 lines and 24.4 tokens on average.
Such a number of generated tokens is far within the maximum number of tokens in recent LLMs (\eg{} 2,048 for \wizardcoder{}~\cite{luo2023wizardcoder}). Therefore, it remains unclear about the further potential of  LLMs in generating long code snippets. 
\textit{Second, such evaluation mainly focuses on generating one single code unit, \eg{} one function or one statement.} However, as shown in previous work~\cite{yu2023codereval}, only 30\% of methods are independent to other code contexts in the open-source projects. Therefore, it remains unclear how LLMs perform in generating a compound code unit of multiple methods~\footnote{As  we currently focus on Python, we distinguish concepts ``\textit{method}'' and ``\textit{function}'': a  method is associated to an object and requires an object instance to be invoked, while a function is an independent code block that can be called from anywhere.} which are dependent to each other (\eg{} invoking each other or accessing a same variable). 

\textbf{Benchmark \ourbenchmark{}}. To fill this knowledge gap, this work makes the first attempt to evaluate LLMs in a more \textit{challenging} code generation scenario, \ie{} class-level code generation. In particular, we evaluate the model capability of generating a class of multiple interdependent methods for the given natural language description. As no existing benchmarks cover class-level code generation tasks, we  manually construct the first class-level code generation benchmark \ourbenchmark{} in a rigorous   and time-intensive way, which takes approximately 500 person-hours to construct 100 class-level Python code generation tasks. Overall, \ourbenchmark{} covers a wide range of topics in practical software development (\eg{} management systems and  game development). Each task is constructed with a test suite of high testing sufficiency (\eg{} 98.2\% and 99.7\% branch-level or statement-level coverage) so as to facilitate reliable correctness checking of the generated code; furthermore, each task is designed to generate a class of multiple methods with diverse dependencies (\eg{} field, method, and library dependencies). 

\textbf{Empirical study}. Based on the new benchmark \ourbenchmark{}, we then perform the first study to evaluate LLMs on class-level code generation. In particular, our experiments include 11 state-of-the-art LLMs, which are diverse in model sizes, foundation models, sources, or domains. For each studied LLM, we explore its performance in generating class-level code with three different generation strategies, \ie{} holistic generation (generating the entire class all at once), incremental generation and compositional generation (generating the class method by method). For each generated code snippet, we measure its correctness with the widely-used metric \passk{}~\cite{chen2021pass_at_k}. In addition, we also investigate the model ability of generating dependent code and analyze  bad cases of incorrect classes.

\textbf{Main findings and implications.} Based on our results, we have the following main findings. First, we find that all existing LLMs show much worse performance on class-level code generation compared to on standalone method-level code generation benchmarks like \humaneval{}; and the method-level coding ability cannot equivalently reflect the class-level coding ability among LLMs. Second, we find that \gptfour{} and \gptthree{} still exhibit dominate superior than other LLMs on class-level code generation, and the second-tier models includes \starcoder{}, \inscodegen{}, and \wizardcoder{} with very similar performance. Third, we find that generating the entire class all at once (\ie{} holistic generation strategy) is the best generation strategy only for \gptfour{} and \gptthree{}, while step-by-step generation (\ie{} incremental and compositional) is better strategies for the other models with limited ability of understanding long instructions and utilizing the middle information. Lastly, we find the limited model ability of generating method-dependent code and discuss the frequent error types in generated classes. 

In summary, this paper makes the following contributions:
\begin{itemize}[leftmargin=15pt, topsep=5pt]
    \item \textbf{The first benchmark \ourbenchmark{}} for class-level code generation, which is manually constructed with 500 person-hours and publicly available on ~\cite{classeval};
    % \yiling{change the link to public one}; 
    
    \item \textbf{The first study} to evaluate 11 representative LLMs on class-level code generation with three different generation strategies;
    
    \item \textbf{Findings and implications} on analyzing the  model capability and future directions for LLMs on class-level code generation. 
\end{itemize}

%% file: sections/background.tex
\section{Background}\label{sec:back}
We first introduce the recent LLMs for code generation in Section~\ref{sec:back:codellm} and then motivate our study by revisiting existing code generation benchmarks in Section~\ref{sec:back:bench}. 

\input{sections/background/codellm}

\input{sections/background/benchmark}

%% file: sections/background/codellm.tex
\subsection{Large Language Models for Code Generation}
\label{sec:back:codellm}

Code generation is a task focusing on generating code snippets for the given natural language description, which has been extensively studied in recent literature~\cite{vikram2023large,10172763,kang2023explainable}. \textit{General LLMs} (\eg{} GPT-4~\cite{openai2023gpt4} and ChatGLM~\cite{du2022chatglm}), which are large models with more than billions of parameters trained on general textual/code corpora and instructions, demonstrate remarkable capabilities not only in general NLP tasks~\cite{chang2023survey} but also promising performance in code generation. For example, GPT-4 achieves the highest pass rate on \humaneval{} benchmark~\cite{luo2023wizardcoder}. Therefore, there has recently been an increasing trend to evaluate the code generation capacity even for general LLMs~\cite{chen2021huamneval,shen2023pangucoder2}. 
\textit{Code LLMs}, which are large models mainly trained with massive code-specific corpora and instructions, often have better capability than general LLMs in code generation tasks~\cite{luo2023wizardcoder, christopoulou2022pangucoder,zan-etal-2023-large}.
Existing code LLMs are designed with different training objectives. For example, some are using next-token prediction, while  some code LLMs (\eg{} InCoder~\cite{fried2022incoder} and StarCoder~\cite{li2023starcoder}) are trained with ``filling-in-the middle'' (FIM) capability, \ie{} infilling the missing portion based on the context. To date, a large number of code LLMs have been proposed, such as WizardCoder~\cite{luo2023wizardcoder}, Instruct-StarCoder~\cite{intructstarcoder}, and Instruct-CodeGen~\cite{intructcodegen}.

%% file: sections/background/benchmark.tex
\input{tables/benchmarks}
\subsection{Existing Benchmarks for Code Generation}
\label{sec:back:bench}
Code generation benchmarks typically include various coding tasks where a natural language description serves as input, and the corresponding code serves as the ground truth output. Evaluation metrics such as passing rate (\passk~\cite{chen2021pass_at_k}) are commonly used to assess the correctness of the generated code.

To date, many code generation benchmarks have been constructed via automated or manual manners. In this study, we revisit widely-used code generation benchmarks from the three following sources: (i) Top-10 popular datasets with the highest download volumes from Huggingface code generation datasets~\cite{codegeneration}, (ii) benchmarks associated with recent LLM papers (released between June 2021 and June 2023), and (iii) enhanced benchmarks such as \humaneval{}+~\cite{liu2023humanevalplus} and Multi-HumanEval~\cite{athiwaratkun2023multilingual}.
Table~\ref{table:benchmark} provides an overview of the 13 distinct benchmarks collected from the three sources, including their release time, construction method (\ie manually written or automatically collected from public code corpus or competitions), benchmark size (\#Tasks), target code granularity, target code language, code scale (\#LOC: average lines of code, \#Tokens: average number of tokens), average number of test cases per task (\#Tests), and detailed input information. We also present our constructed benchmark \ourbenchmark{} in the last row for comparison.

\textbf{Based on Table~\ref{table:benchmark}, we find that existing benchmarks actually shape a rather \textit{simple} code generation scenario, which mainly evaluate the capability of LLMs in generating one single code unit (a function or a statement) in a rather \textit{standalone} way.} 
In particular, existing benchmarks typically focus on function-level or statement-level code generation tasks (Column ``Granularity'') and rarely include additional code contexts in the input (Column ``Input Information''), which assumes that the code to be generated is an independent unit and thus leads to two limitations in evaluating LLMs.

First, existing benchmarks mainly focus on short code generation tasks, like generating one function or one statement. These tasks typically involve limited number of lines (\eg 1 to 30) and tokens (\eg 4.6 to 108.2), which may not fully explore the capabilities of recent LLMs that can handle much longer sequences, such as \wizardcoder{} with 2,048 tokens. Thus, the potential of  LLMs in generating longer code snippets remains unclear.
Second, existing benchmarks mainly focus on generating independent code units without considering other code contexts. For instance, as shown in Figure~\ref{fig:benchmark_example}, benchmarks like \mbpp{} and \humaneval{} only provide limited information as input, such as natural language descriptions or function signatures with example inputs and outputs. However, in real-world scenarios, methods often depend on each other or share variables. 
Previous work~\cite{yu2023codereval} indicates that only 30\% of methods in open-source projects are independent of other code contexts.
Therefore, it remains unclear how LLMs perform in generating a compound code unit of multiple methods which are dependent to each other (\eg{} invoking each other or accessing a same variable).

\input{image/benchmark_example}

\textbf{Our Motivation.} Existing benchmarks cannot facilitate the model evaluation on more complicated code generation tasks, such as generating longer and compound code units of multiple interdependent methods. To address this gap, we manually construct the first class-level code generation benchmark \ourbenchmark{} and perform the first study to evaluate LLMs on class-level code generation tasks, which ask LLMs to generate a class of multiple interdependent methods based on a given natural language description.

%% file: tables/benchmarks.tex
\begin{table*}[htb]
	\centering
 % \vspace{-1mm}
	\caption{Existing Benchmarks for Code Generation }
   % \vspace{-1mm}
    \label{table:benchmark}
	\begin{adjustbox}{width=2.1\columnwidth}
	\begin{tabular}{l|c|c|m{3cm}<{\centering}|c|c|c|c|c|c|m{7cm}<{\centering}}
		\hline
        \textbf{Benchmark} & \textbf{Time}  & \textbf{Language} &  \textbf{Manual/Automated} &  \textbf{Source} & \textbf{Granularity} & \textbf{\#Tasks}  & \textbf{\#Tests}  & \textbf{\#LOC}  & \textbf{\#Tokens} & \textbf{Input Information} \\ \hline

        Concode~\cite{iyer2018concode} & 2018 & Java & Automated & Github & Function-level & 2,000 & - & - & 26.3 & NL\\ \hline
        CoNaLA~\cite{yin2018conala} & 2018 & Python & Automated & Stack Overflow & Statement-level & 500 & - & 1 & 4.6  & NL\\ \hline
        APPS~\cite{hendrycks2021apps} & 2021 & Python & Automated & Contest Sites & Competitive & 5,000 & 13.2 & 21.4 & 58  & NL + Example Inputs/Outputs\\ \hline
        \humaneval{}~\cite{chen2021huamneval} & 2021 & Python & Manual & -& Function-level & 164 & 7.7 & 11.5 & 24.4 &NL + Function Signature + Example Inputs/Outputs \\ \hline
        MBPP~\cite{austin2021mbpp} & 2021 & Python & Manual & -& Function-level & 974& 3.0 & 6.8 & 24.2 & NL\\ \hline
        math-qa~\cite{austin2021mbpp} & 2021 & Python & Manual & Math Study Sites & Statement-level & 2,985 & - & 7.6 & 24.6 & NL\\ \hline        
        Multi-\humaneval~\cite{athiwaratkun2023multilingual} & 2022 & Multilingual & Manual & - & Function-level & 164 & 7.7 & 11.5  & 24.4 & NL + Function Signature + Example Inputs/Outputs\\ \hline
        MBXP~\cite{athiwaratkun2023multilingual} & 2022 & Multilingual & Manual & - & Function-level & 974 & 3.0 & 6.8 & 24.2 & NL\\ \hline
        multi-math-qa~\cite{athiwaratkun2023multilingual} & 2022 & Multilingual & Manual & Math Study Sites & Statement-level & 2,985 & - & 7.6 & 24.6 & NL\\ \hline        
        CodeContests~\cite{li2022codecontest} & 2022 & Python, C++ & Automated & Contest Sites & Competitive & 165 & 203.7 & 59.8 & 184.8  & NL + Example Inputs/Outputs\\ \hline
        DS-1000~\cite{lai2023ds1000} & 2022 & Python & Automated & Stack Overflow & Statement-level & 1,000 & 1.6 & 3.8 & 12.8   & NL\\ \hline
        \humaneval{}+~\cite{liu2023humanevalplus} & 2023 & Python & Manual & - & Function-level & 164 & 774.8 & 11.5 & 24.4 & NL + Function Signature + Example Inputs/Outputs\\ \hline        
        CoderEval~\cite{yu2023codereval} & 2023 & Python, Java & Automated & Github & Function-level & 230 & - & 30 & 108.2 & NL + Function Signature\\ \hline \hline

        \textbf{\ourbenchmark{}} & \textbf{2023} & \textbf{Python} &  \textbf{Manual} & - & \textbf{Class-level} & \textbf{100} & \textbf{33.1} & \textbf{45.7} & \textbf{123.7} & \textbf{Class Skeleton}\\ \hline 

	\end{tabular}
	\end{adjustbox}
\end{table*}

%% file: image/benchmark_example.tex
\begin{figure}[htb]
	\centering
        \vspace{-1mm}
	\includegraphics[width=0.9\columnwidth]{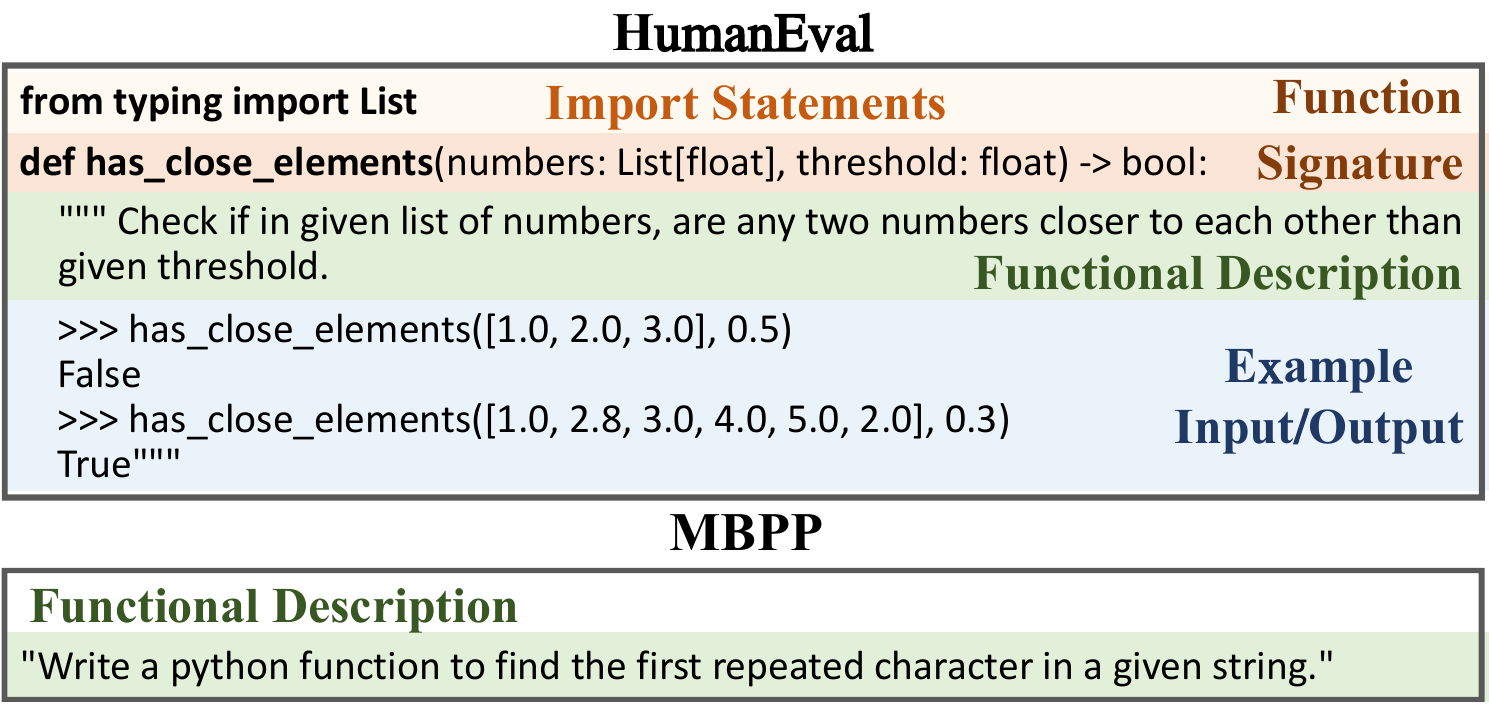}
	\caption{Examples in Existing Benchmarks}
	\label{fig:benchmark_example}
        \vspace{-1mm}
\end{figure}

%% file: sections/approach.tex
\section{New benchmark \ourbenchmark}
\label{sec:bechmark}
In this section, we introduce our new benchmark \ourbenchmark. We present the benchmark format (Section~\ref{sec:benchmark:format}), the  construction procedure (Section~\ref{sec:benchmark:construction}), and the benchmark characteristics (Section~\ref{sec:benchmark:result}).

\input{sections/benchmark/format}

\input{sections/benchmark/construction}

\input{sections/benchmark/resulted_benchmark}

%% file: sections/benchmark/format.tex
\subsection{Benchmark Format}\label{sec:benchmark:format}
Each coding task in \ourbenchmark{} comprises an input description for the target class (\ie the class to be generated), a test suite for verifying the correctness of the generated code, and a \textit{canonical solution} that acts as a reference implementation of the target class.

\input{image/classeval_example}
Typically, LLMs generate code snippets based on input descriptions and the  correctness is verified with the provided test suite.
The generated code must conform to a consistent interface (\eg{} the types of input parameters and return values) specified in the test suite for valid execution. 
For example, the benchmark \humaneval{} specifies the signature of the target function (Figure~\ref{fig:benchmark_example}) to ensure that the generated bodies are validly checked by the given test suite.
To achieve this, we define a \textbf{\textit{class skeleton}} format for the input descriptions in our coding tasks. The class skeleton serves as a structured blueprint for the target class, containing both class-level information (import statements, class name, class description, and class constructor) and method-level information (method signature, functional description, parameter/return descriptions, and example input/outputs). 
The detailed definitions of elements in the class skeleton are in Table~\ref{table:skeleton}. Column ``Mand.'' indicates whether the element is mandatory in the class skeleton.
Method-level elements are all adopted from existing benchmarks like \humaneval{}. 
Figure~\ref{fig:sdeval_example} further illustrates an example of a class skeleton, with different components highlighted in various colors.
The class skeleton, inspired by contract programming~\cite{contract}, serves as formal and precise specifications for code generation by outlining expected behaviors, pre-conditions, and post-conditions. LLMs generate class-level code that aligns with the given test suite based on the class skeleton.

\input{tables/skeleton}

%% file: image/classeval_example.tex
\begin{figure}[htb]
	\centering
        %\vspace{-4mm}
	\includegraphics[width=0.88\columnwidth]{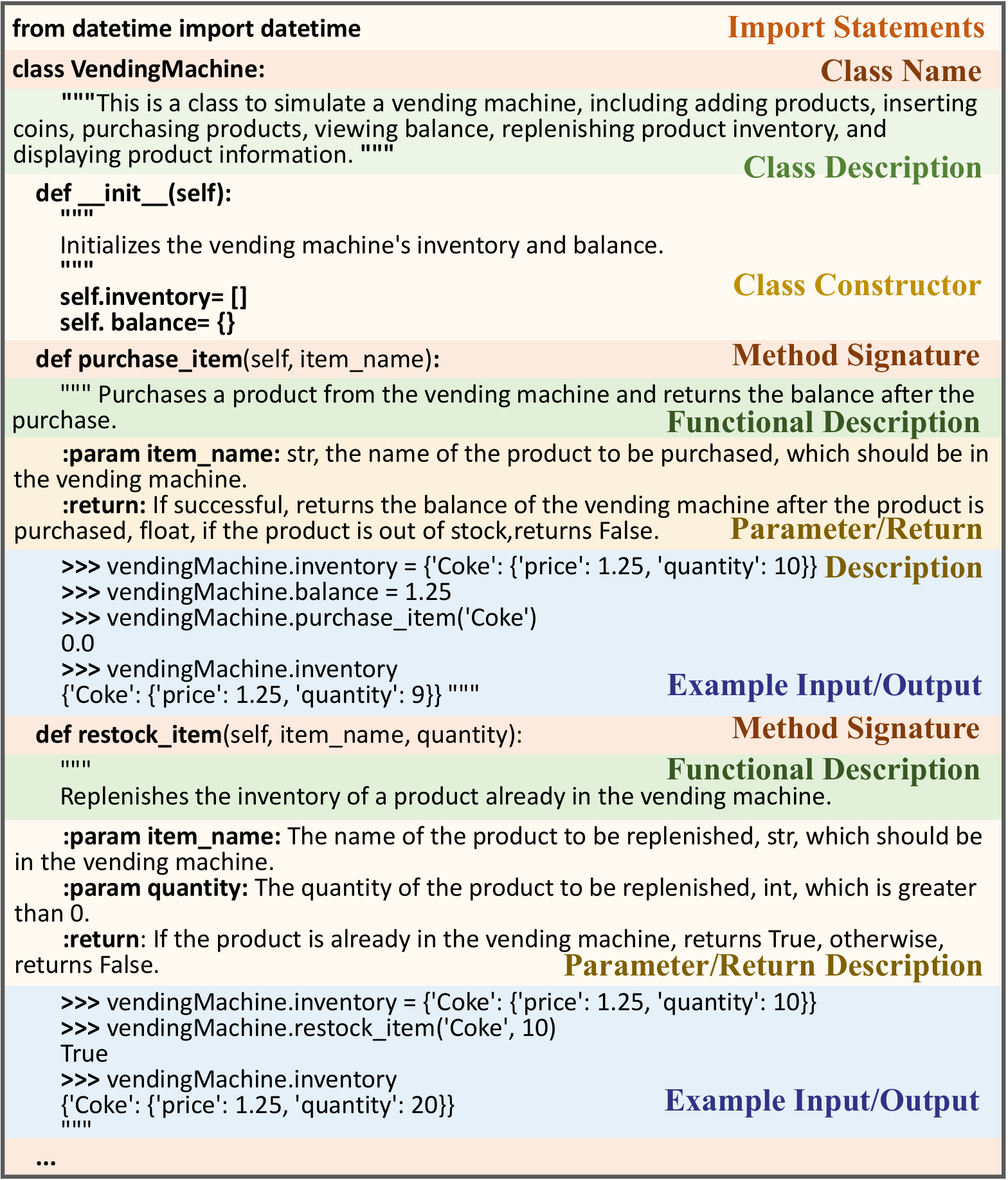}
	\caption{An Example of Class Skeleton in \ourbenchmark}
        % \todo{update the Figure according to the text}}
	\label{fig:sdeval_example}
        %\vspace{-2mm}
\end{figure}

%% file: tables/skeleton.tex
\begin{table*}[htp]
	% \footnotesize
	\centering
    \vspace{-3mm}
    \caption{Elements Defined in Class Skeleton}
	\label{table:skeleton}
	\vspace{-2mm}
        \begin{adjustbox}{width=2\columnwidth}
	\begin{tabular}{l|l|c|l}
		\hline

\multicolumn{2}{l|}{\textbf{Elements}} & \textbf{Mand.} & \textbf{Definition} \\ \hline
\multirow{4}{*}{\tabincell{l}{Class \\ Level \\ Info.}} & 
Class Name  & \checkmark & The name of the target class \\

& Class Description & \checkmark &  The description of the overall functionality of the target class\\

&Import Statements & \ding{53}  &  Indicating the external libraries or modules necessary for implementing the target class \\

& Class Constructor & \ding{53} &  The initial method automatically invoked to initialize the attributes once the class is instantiated \\ \hline

\multirow{4}{*}{\tabincell{l}{Method  \\ Contract \\ Design}} &  Method Signature & \checkmark &  Defining the target method name, input parameters, and return type\\ 

& {Functional Description} & \checkmark &  Natural language descriptions on the functionality of each method \\

& {Parameter/Return Description} & \ding{53} &  Textual descriptions on expected inputs (\eg{} parameter types) and outputs (\eg{} return values) for each method \\

& {Example Input/Output} & \ding{53} &  Concrete examples of  input values and corresponding output values on executing the target method\\

\hline
	\end{tabular}
        \end{adjustbox}
	% \vspace{-7.5mm}
\end{table*}

%% file: sections/benchmark/construction.tex
\subsection{Benchmark Construction Procedure}
\label{sec:benchmark:construction}
Figure~\ref{fig:overview} illustrates the procedure of constructing \ourbenchmark{}. 
We follow four steps to create \ourbenchmark{}: (i) select suitable coding  tasks using different strategies (Section~\ref{sec:task_selection}); (ii) construct class skeletons based on the principles of contract programming~\cite{contract} and test-driven development~\cite{bhat2006testdriven} (Section~\ref{sec:benchmark:skeleton}); (iii) create the test suite for each class skeleton (Section~\ref{sec:benchmark:test}); and (iv) write the canonical solution for each coding task (Section~\ref{sec:benchmark:solution}). The  constructed class skeletons, test suites, and canonical solutions form our class-level code generation benchmark \ourbenchmark{}.

\input{image/construction}

To avoid the coding tasks being seen by LLMs during their training, our benchmark is constructed \textit{completely manually}, so as to mitigate potential data leakages from existing code sources. Our manual construction involves a time-intensive process with approximately 500 person-hours on constructing 100 class-level coding tasks. Due to the significant manual efforts required, we currently stop the benchmark scale to this size.
Moreover, following the trend of most existing benchmarks~\cite{chen2021huamneval, austin2021mbpp}, our benchmark primarily focuses on Python given its prevalence~\cite{srinath2017python}.

\input{sections/benchmark/task_selection}

\input{sections/benchmark/skeleton_construct}
\input{sections/benchmark/test_construct}
\input{sections/benchmark/solution_construct}

%% file: image/construction.tex
\begin{figure}[htb]
	\centering
        %\vspace{-2mm}
	\includegraphics[width=0.87\columnwidth]{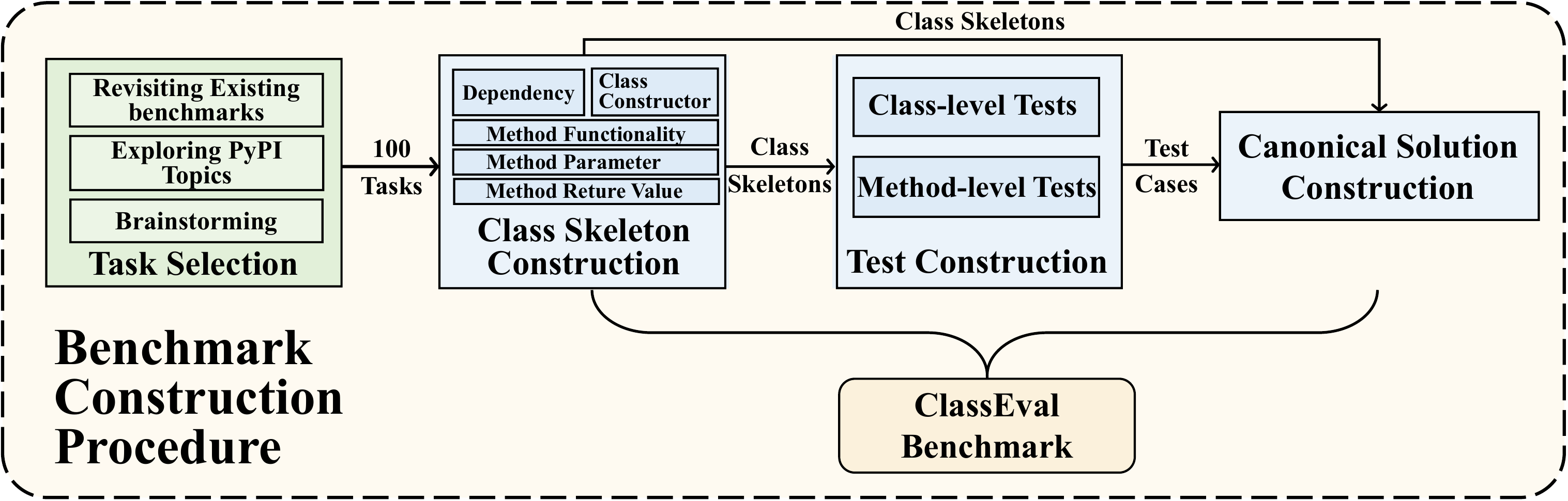}
	\caption{Overview of \ourbenchmark{} Construction Process}
	\label{fig:overview}
        %\vspace{-2mm}
\end{figure}

%% file: sections/benchmark/task_selection.tex
\vspace{-3mm}
\subsubsection{Task Selection}~\label{sec:task_selection}
In this step, we design class-level coding tasks (\ie{} a unique \textit{class description} for each task as defined in Table~\ref{table:skeleton}) for our benchmark. 

\textbf{Inclusion Sources.} 
We design our coding tasks to cover diverse and real-world development topics, based on the following three sources.
(i) \textit{Revisiting Existing Benchmarks.} We refer to well-established benchmarks like \humaneval and \mbpp (Table~\ref{table:benchmark}) to include prevalent and common topics.
(ii) \textit{Exploring PyPI Topics}. We manually explore the Python Package Index (PyPI)~\cite{pypi}, which hosts a vast repository of Python software packages and provides a diverse range of potential task topics.
(iii) \textit{Brainstorming}. 
All authors (with 2-8 years of Python development experience) actively participate in brainstorming to generate potential coding tasks beyond the ones collected above.

\textbf{Exclusion Criteria.} Our benchmark focuses on coding tasks that can be implemented within one single class. Therefore, we exclude tasks that have complicated dependencies on the execution environment, including those related to (i) Network Programming, (ii) Graphical User Interface (GUI) Design, (iii) Data Visualization, (iv) System Programming, and (v) Concurrent Programming. These tasks often require interactions with other classes or cannot be easily verified with assertion statements in unit tests.

In this way, we obtain a list of 100 diverse class-level coding tasks, covering a wide spectrum of topics, such as \topicgame{}, \topicfile{}, and \topicproject{}. Table~\ref{table:topic} presents the topic distribution of our tasks.
\input{tables/topic}

%% file: tables/topic.tex
\begin{table*}[htp]
	\footnotesize
	\centering
 \vspace{-1mm}
	\caption{Topic Type Definitions in \ourbenchmark{}} 
	\label{table:topic}
	\vspace{-2mm}
        \begin{adjustbox}{width=2.1\columnwidth}
	\begin{tabular}{cm{9cm}<{\raggedright}m{5.5cm}<{\raggedright}c}
		\hline
		\textbf{Topic} & \textbf{Description} & \textbf{Examples} & \textbf{\#Tasks} \\ \hline
            \topicproject{} & Operational functionalities in common software management systems projects & Student Registration System, Movie Booking System & 27 \\ \hline
            \topicdataprocess{} & Processing data according to specific rules or patterns & Text-to-number Conversion, URL Format Validation & 26 \\ \hline
            \topicmath{} & Algorithms for mathematical and statistical problems & Basic Arithmetic Operations, Area Calculation & 16 \\ \hline
            \topicgame{} & Algorithms for game functionalities, including mechanics and state management & Minesweeper Game, Gomoku Game & 10  \\ \hline            
            \topicfile{} & Common file operations including reading, writing, and simple processing data in files &  CSV File Processor, JSON File Processor & 9 \\ \hline
            \topicdatabase{} & Implementation of common database operations & Library Database Operation, SQL Query Generator & 7 \\ \hline
            \topicnlp{} & Techniques for processing and analyzing text data & Stop Word Removal, Longest Word Identification & 5 \\ \hline
	\end{tabular}
        \end{adjustbox}
	% \vspace{-7.5mm}
\end{table*}

%% file: sections/benchmark/skeleton_construct.tex
\subsubsection{Class Skeleton Construction} 
\label{sec:benchmark:skeleton}
During this step, we manually construct the class skeleton for each coding task, involving 5 participants with an average of 3 years of Python development experience. 
Each skeleton is initially assigned to two participants, one responsible for writing the class skeleton and the other for double-checking it. In case of disagreements, a third participant facilitates discussions to reach a consensus on the class skeleton. The entire process adheres to the following design principles.

\textbf{Principle 1 (dependency):} Each class skeleton should contain methods with diverse dependencies, \ie{} the methods are dependent to other code contexts within the class.
Previous work~\cite{yu2023codereval} has shown that the majority of methods (over 70\%) are dependent on other code contexts in the project.
Unlike previous benchmarks (\eg{} \humaneval{} and \mbpp shown in Figure~\ref{fig:benchmark_example}) that focus on standalone function-level code generation,
our class-level benchmark aims to capture the real-world scenario where methods often have dependencies with other code contexts.
To distinguish our benchmark from function-level ones, we deliberately avoid tasks that generate a class with independent methods, which would essentially be a collection of individual method-level coding tasks. Instead, class skeletons in our benchmark includes methods with diverse dependencies, including (i) \textit{Library Dependency}, where methods rely on external libraries; (ii) \textit{Field Dependency}, where methods depend on class instance variables (fields); (iii) \textit{Method Dependency}, where methods rely on other methods within the same class; and (iv) \textit{Standalone}, where methods function independently without dependencies on fields, methods, or external libraries.

\textbf{Principle 2 (class constructor):} The class constructor (if has) in each class skeleton should define the class fields and their default values. The constructor also includes natural language descriptions of the class fields to provide a clear understanding of their meanings. Importantly, the constructor does not make calls to other methods within the class to preserve the independence and self-contained nature of the class initialization process. 

\textbf{Principle 3 (method functionality):} 
We avoid including complex functionalities like  
closing database connections, which are not easily testable and verifiable. Additionally, we enhance code reusability and maintainability by breaking down common and repetitive functionalities into separate methods. This principle fosters potential interdependencies between methods, simulating a more interconnected and practical coding scenario.

\textbf{Principle 4 (method parameter):} The method parameters are limited to primitive data types, avoiding object-level parameters or loosely defined arguments like \textit{**kwargs}. This principle not only enhances clarity in method invocation but also facilitates testing, making it easier to create unit tests and verify the functionality of individual methods in isolation.

\textbf{Principle 5 (method return value):} Methods should include return values whenever possible for testing. For indicating success or failure, they use Boolean return types for standardization instead of custom strings. Additionally, method designs may encompass evaluative conditions for input parameters and include exception handling mechanisms. Detailed specifications of exception types, message content, and triggering circumstances are provided to ensure comprehensive testing and validation of exception handling.

Each constructed class skeleton would contain mandatory elements (\ie{} the class description, the class name, the method signature, and the functional description) and optional elements (\ie{} import statements, class constructor, parameter/return descriptions and the example input/output). 

%% file: sections/benchmark/test_construct.tex
\subsubsection{Test Construction}\label{sec:benchmark:test}
In this step, we manually construct a test suite for each coding task based on its class skeleton. The participants who were responsible for creating the class skeleton now take on the task of writing the corresponding test suite. Similarly, one participant focuses on writing the unit test cases, while the other ensures the quality and correctness of the test cases.

The methods in each class skeleton are designed to have multiple dependent relationships, as mentioned in Principle 1 in Section~\ref{sec:benchmark:skeleton}. Therefore, participants are required to construct test cases at two levels: \textit{method-level tests} and \textit{class-level tests}, so as to fully test the correctness of the implemented methods when they are invoked individually or together. 
\textit{Method-level tests} primarily check the correctness of each method under test by independently invoking it without invoking any other methods in the class. On the other hand, \textit{class-level tests} mainly check the correctness of multiple methods under test by  invoking them sequentially together. Method-level tests ensure that the correctness of each method under test is individually checked without being impacted by the incorrect implementation of other methods, while class-level tests evaluate the overall correctness of the class by considering its interactions. Figure~\ref{fig:test_example} provides two examples of both method-level and class-level test cases constructed for the class skeleton in Figure~\ref{fig:sdeval_example}. Additionally, we include examples of test cases from existing benchmarks \humaneval and \mbpp to highlight the differences. The function-level tests in existing benchmarks are comparable to the method-level tests in our benchmark, but the major difference is that function-level tests in existing benchmarks only check the return values of the function under test while our method-level tests further check the fields of the class. As shown in Figure \ref{fig:test_example}, when testing the \texttt{purchase\_item} method, the method-level test in \ourbenchmark{} not only verifies the return value but also evaluates the operations performed on the \texttt{inventory} field. Moreover, existing benchmarks lack class-level tests since they primarily focus on single-function generation.

We then introduce the main principles of constructing method-level tests and class-level tests, respectively. For method-level tests, participants are asked to create at least five test cases to cover diverse scenarios of each method under test. For class-level tests, participants are required to construct test cases with different combinations of methods under test, ensuring that each method is invoked at least once in the class-level tests. To simplify test construction, participants are required to use the existing unittest framework~\cite{unittest}, which provides diverse assertion APIs and a set of Test Fixtures (\eg \texttt{setUp} and \texttt{tearDown} methods) for preparation and cleanup tasks before and after test execution. Additionally, all constructed test cases are limited to a five-second running time to prevent potential infinite loops in the generated code.

\input{image/test}

%% file: image/test.tex
\begin{figure*}[htb]
	\centering
    \vspace{-2mm}
	\includegraphics[width=2.0\columnwidth]{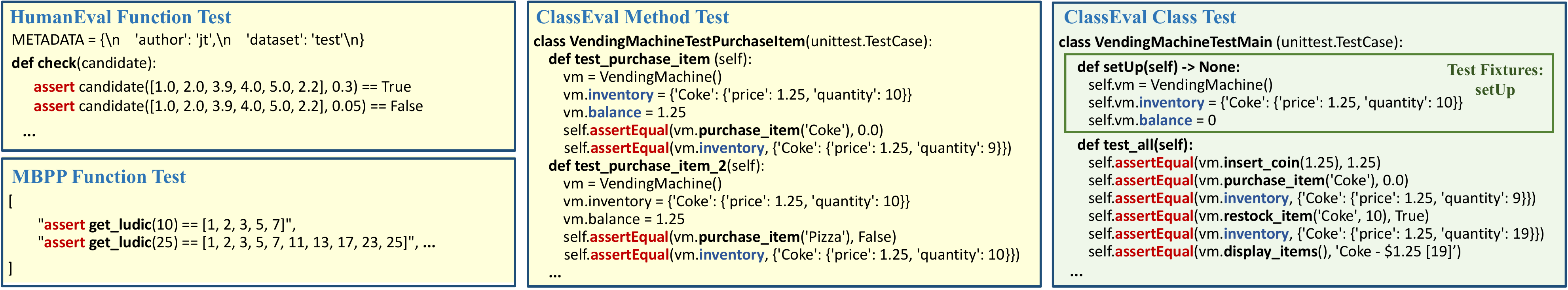}
    \vspace{-1mm}
	\caption{Test Cases in Existing Benchmarks and \ourbenchmark{}} 
	\label{fig:test_example}
        %\vspace{-4mm}
\end{figure*}

%% file: sections/benchmark/solution_construct.tex
\subsubsection{Canonical Solution Construction}\label{sec:benchmark:solution}
In this step, we manually write the canonical solution for each coding task based on its constructed class skeleton and test cases. Four participants (each with 2 - 4 years of Python development experience) who were not involved in constructing the class skeletons and test cases are engaged in this step. Each coding task is assigned to two participants, with one responsible for writing the canonical solution and the other for double-checking it. Participants are required to execute the solutions with test cases to identify and fix any bugs.

%% file: sections/benchmark/resulted_benchmark.tex
\subsection{Benchmark Characteristics}
\label{sec:benchmark:result}

In this way, we manually build a new benchmark \ourbenchmark{} of 100 class-level coding tasks.
The detailed characteristics are as follows.

\textbf{Scale.} \ourbenchmark{} consists of 100 classes and 412 methods. To facilitate a direct comparison with other code generation benchmarks, we include the statistical data of \ourbenchmark{} in Table~\ref{table:benchmark}. The results reveal large differences in lines of code for \ourbenchmark{} (45.7) compared to the two most widely used handwritten benchmarks, \humaneval{} and \mbpp{}, with multipliers of 4.0 and 6.7 respectively. Additionally, we perform additional statistics on the average number of tokens in the entire docstring information (class skeleton) in \ourbenchmark{} (259.3), surpassing \humaneval (67.7) and MBPP (14.5) by a factor of 3.8 and 17.9 respectively. These results demonstrate that the class-level code generation task in \ourbenchmark{} presents higher complexities, involving longer code generation, as well as more detailed and sophisticated docstring information.

\input{tables/test}
\textbf{Test Sufficiency.}
Table~\ref{table:test} provides coverage statistics for the test cases in our benchmark compared to \humaneval{} and \mbpp{}. We collect the statement-level and branch-level coverage of the test cases on the canonical solution code using the Python toolkit \textit{coverage}~\cite{coverage}. Additionally, we provide the average number of method-level tests (\#Tests/M) and average class-level tests (\#Tests/C). As shown in Table~\ref{table:test}, the test cases in \ourbenchmark{} achieve significantly higher statement-level and branch-level coverage (both over 98\%) compared to \humaneval{} and \mbpp{}. This indicates more extensive code checking for the generated solutions in our benchmark, which is supported by the fact that \ourbenchmark{} also includes a larger number of method-level and class-level tests on average.

\textbf{Dependency.} \ourbenchmark{} focuses on class-level code generation tasks, distinguishing it from previous benchmarks. Table~\ref{table:dependency} shows the distribution of dependency levels within methods across \ourbenchmark{} and previous benchmarks, as explained in Section~\ref{sec:benchmark:format}. Notably, Library, Field, and Method dependencies are not mutually exclusive, and some methods may have a combination of Field and Method dependencies. We classify methods with either Field or Method dependencies as class-level dependent methods, totaling 314 (76.2\%) within \ourbenchmark{}. 
This inclusion makes \ourbenchmark{} a comprehensive benchmark, suitable for evaluating LLMs that must account for intricate class-level interactions and contextual dependencies.

\input{tables/dependency}

Overall, in comparison to previous manually-crafted code generation benchmarks, \ourbenchmark{} contains complicated class-level coding tasks involving larger-scale code snippets, diverse dependencies, sufficient test cases, and a wider range of topics from practical software development.

%% file: tables/test.tex
% \vspace{-2mm}
\begin{table}[htp]
	\footnotesize
	\centering
	\caption{Test Coverage and Test Cases Statistics}
	\label{table:test}
	\vspace{-2mm}
	\begin{tabular}{ccccc}
		\hline
		\textbf{Benchmark} & \textbf{Statement} & \textbf{Branch} & \textbf{\#Tests/M} & \textbf{\#Tests/C} \\ \hline
		\humaneval{}  & 98.8\% & 83.2\% & 7.7 & -    \\ \hline
		\mbpp{} & 98.6\% & 76.4\% & 3.0 & -     \\ \hline
        \ourbenchmark{} & 99.7\% & 98.2\% & 8.0 & 33.1 \\ \hline
	\end{tabular}
	\vspace{-1mm}
\end{table}

%% file: tables/dependency.tex
\begin{table}[htp]
	\footnotesize
	\centering
	\caption{Comparative Distribution of Dependency Levels}
    \vspace{-1mm}

	\label{table:dependency}
	\begin{tabular}{cccc}
		\hline
		\textbf{Dependency} & \textbf{\mbpp{}} & \textbf{\humaneval{}} & \textbf{\ourbenchmark{}}\\ \hline
	Standalone  & 974 (100\%) & 157 (95.8\%) & 58 (14.1\%)   \\ \hline
	Library   & - & 7 (4.2\%)  &  89 (21.7\%)   \\ \hline
        Field & - & - &  269 (65.5\%)    \\ \hline 
        Method & - & -  &  107 (26.0\%)   \\ \hline
	\end{tabular}
\end{table}

%% file: sections/evaluation/evaluation.tex
\section{Empirical Study}
Using \ourbenchmark, we conduct the first study to evaluate existing LLMs on class-level code generation by answering the following research questions.

\begin{itemize}[leftmargin=15pt]
\item \textbf{RQ1 (Overall Correctness):} how do LLMs perform on class-level code generation?
\item \textbf{RQ2 (Generation Strategies):} how do different generation strategies perform for LLMs on class-level code generation?
\item \textbf{RQ3 (Dependency Generation):} how do LLMs perform on generating code dependent to other contexts during class-level code generation?
\item \textbf{RQ4 (Bad Case Analysis):} what are the common errors during  class-level code generation?
\end{itemize}

\input{sections/evaluation/baselines}

\input{sections/evaluation/class_gen_app}
\input{sections/evaluation/metrics}
\input{sections/evaluation/setting}

\input{sections/results}

%% file: sections/evaluation/baselines.tex
\input{tables/baseline}
\subsection{Studied LLMs}
We select the state-of-the-art LLMs that have been widely studied in recent code generation work~\cite{luo2023wizardcoder, liu2023humanevalplus}. In particular, we focus on recent models released since 2022, and we exclude the small models (with less than 1B parameters) due to their limited efficacy or the large models (with more than 20B parameters) due to our resource limits. Table~\ref{table:baseline_info} presents the 11 LLMs studied in our experiments with their releasing time (Column ``Time''), model sizes (Column ``Size''), and base models. In addition, we also summarize the training characteristics of the studied models, including whether the model has been trained to possess the ability of ``filling-in-the-middle'' (FIM) and whether it possesses the instruction-following (IF) ability via instruction tuning. Both FIM and IF capabilities are essential for the class-level code generation tasks.  
As shown in Table~\ref{table:baseline_info}, our study includes a wide scope of LLMs that are diverse in multiple dimensions, such as (i) being both closed-source and open-source, (ii) utilizing different base models, (iii) covering a range of model sizes from 1B to 16B, (iv) being trained by both general or code-specific instructions, and (v) exhibiting different FIM and IF capabilities.

%% file: tables/baseline.tex
\begin{table}[htp]
	% \footnotesize
	\centering
    \caption{Studied LLMs}
	\label{table:baseline_info}
	\vspace{-2mm}
        \begin{adjustbox}{width=\columnwidth}
	\begin{tabular}{c|cccccc}
		\hline
		\multicolumn{2}{c}{\textbf{Model}}  & \textbf{Base Model} & \textbf{Time}  & \textbf{Size} & \textbf{IF} & \textbf{FIM} \\ \hline
        \multirow{7}{*}{\tabincell{c}{Code \\ LLM}} & Instruct-CodeGen~\cite{intructcodegen} & CodeGen-multi~\cite{nijkamp2022condegen} & 2022.3 & 16B & \checkmark & \checkmark    \\ \cline{2-7}
       % &  Instruct-CodeT5+~\cite{wang2023codet5plus} & CodeT5+~\cite{wang2023codet5plus} & 2023.5 &  16B & \checkmark & \checkmark   \\ \cline{2-7}
       & WizardCoder~\cite{luo2023wizardcoder} & StarCoder~\cite{li2023starcoder} & 2023.6 & 15B & \checkmark & \checkmark    \\ \cline{2-7}
        & Instruct-StarCoder~\cite{intructstarcoder} & StarCoder~\cite{li2023starcoder} & 2023.5 & 15B  & \checkmark  & \checkmark    \\ \cline{2-7}
        & CodeGeeX~\cite{zheng2023codegeex} & - & 2023.3 & 13B  & \ding{53} & \ding{53}     \\\cline{2-7}
        & InCoder~\cite{fried2022incoder} & Dense~\cite{dense} & 2022.4 & 6B & \ding{53} & \checkmark    \\ \cline{2-7}
        % GPT-J~\cite{gpt-j} & 6B & code-specific & \ding{53} & \ding{53}     \\ \hline
        & PolyCoder~\cite{xu2022polycoder} & GPT-2~\cite{radford2019gpt2} & 2022.2 & 2.7B  & \ding{53} & \ding{53}     \\ \cline{2-7}
        % GPT-Neo~\cite{gpt-neo} & 2.7B & code-specific & \ding{53} & \ding{53}     \\ \hline
        & SantaCoder~\cite{allal2023santacoder} & GPT-2~\cite{radford2019gpt2} & 2023.1 & 1.1B  & \ding{53} & \checkmark    \\ \hline
        \multirow{4}{*}{\tabincell{c}{General \\ LLM}} & 
	Vicuna~\cite{zheng2023vicuna} & LLaMA~\cite{touvron2023llama} & 2023.3 & 7B  &  \checkmark & \checkmark    \\ \cline{2-7}
	& ChatGLM~\cite{du2022chatglm} & GLM~\cite{zeng2022glm130b} & 2022.3 & 6B & \checkmark & \checkmark   \\ \cline{2-7}

        & GPT-3.5~\cite{openai2023gpt4} & - & 2022.11 & -  & \checkmark & \checkmark     \\ \cline{2-7}
        & GPT-4~\cite{openai2023gpt4} & - & 2023.3 & -  & \checkmark & \checkmark     \\ \hline
	\end{tabular}
        \end{adjustbox}
	% \vspace{-7.5mm}
\end{table}

%% file: sections/evaluation/class_gen_app.tex
\subsection{Studied Generation Strategies} 
\label{sec:eval:procedure}
Given a class-level code generation task, we study the performance of each model with three different generation strategies as follows:

\begin{itemize}[leftmargin=15pt]
\item \textbf{\holisticgen{} Generation}: the model is asked to generate the entire class all at once with the class skeleton as inputs. 

\item  \textbf{\incrementalgen{} Generation}: the model is asked to generate the class in a method-by-method manner. Each iteration is based on the method bodies that have been generated in previous iterations. The iterative process repeats until all methods in the class are generated.  

\item  \textbf{\individualgen{} Generation}: the model is asked to generate the class in a method-by-method manner. Each iteration is independent, without considering the other generated methods. All the generated methods are assembled to form the class lastly.
\end{itemize}

The holistic generation strategy evaluates the model ability of handling long and complicated coding tasks all at once, while the incremental and compositional generation strategies focus on step-by-step class completion. The incremental strategy simulates progressive software development, where developers incrementally implement current methods based on existing ones. In constrast, the compositional strategy simulates real-world programming scenarios, where developers implement current methods based on other available method signatures.
The compositional generation strategy is not influenced by the hints (if the implemented methods are correct) or the misleading information (if the implemented methods are incorrect) since it does not use other method  implementation as input.
Notably, both incremental and compositional generation strategies differ from standalone function-level code generation tasks in existing benchmarks like \humaneval{}, since our inputs include the class-level context such as the class constructor and other method signatures in the class skeleton.

\subsection{Prompt Design}
We then describe how we prompt LLMs to solve each class-level code generation task in \ourbenchmark{} with each generation strategy.

\parabf{LLMs with IF ability.} Following the common practice of prompting LLMs with IF ability like \wizardcoder~\cite{luo2023wizardcoder}, we set their prompts of two parts: (i) a \textit{system prompt} as the beginning sentence to initialize the model, and followed by (ii) a \textit{task instruction} to describe the goal of the task. Each generation strategy is set with its specific  \textit{task instruction}, \ie{} \textit{Instruction-H} for holistic generation, \textit{Instruction-I} for incremental generation, and \textit{Instruction-C} for a compositional generation. The prompt template is as follows, and each element is previously defined in Table~\ref{table:skeleton}.

\begin{mdframed}
[linecolor=white!50,linewidth=2pt,roundcorner=10pt,backgroundcolor=myyellow!20]
\noindent \textbf{System Prompt:} Provided below is an instruction detailing a task. Compose a response that aptly fulfills the request.
\end{mdframed}

\begin{mdframed}
[linecolor=white!50,linewidth=2pt,roundcorner=10pt,backgroundcolor=myyellow!20]
\noindent \textbf{Instruction-H:} Please complete the class \$\{Class Name\} in the subsequent code. \$\{Class Skeleton\}
\end{mdframed}

\begin{mdframed}
[linecolor=white!50,linewidth=2pt,roundcorner=10pt,backgroundcolor=myyellow!20]
\textbf{Instruction-I:} Please complete the method \$\{Method Name\} within the following class \$\{Class Name\}. \$\{Class-level Info\} \$\{Generated Methods with Contract Designs\} \$\{Target Method Contract Design\}
\end{mdframed}

\begin{mdframed}
[linecolor=white!50,linewidth=2pt,roundcorner=10pt,backgroundcolor=myyellow!20,leftmargin=2cm]
\textbf{Instruction-C:} Please complete the method \$\{Method Name\} within the following class \$\{Class Name\}. \$\{Class-level Info\} \$\{Other Method Signatures\} \$\{Target Method Contract Design\}
\end{mdframed}

\parabf{LLMs without IF ability.}
The prompt of these models is the code context without any instruction: (i) for holistic generation, the prompt is just the class skeleton; (ii) for incremental generation, the prompt in each iteration includes  the class-level information, generated methods, and the target method contract design; (iii) for compositional generation, the prompt for each method includes the class-level information, other method signatures, and the target method contract design. 

%% file: sections/evaluation/metrics.tex
\subsection{Metrics}\label{sec:metrics}
For correctness evaluation, we use the widely-used \passk~\cite{chen2021pass_at_k} metric, which  calculates the percentage of problems solved based on $k$ code samples generated for each task:

\vspace{-4mm}
\begin{equation}
\label{eq:passk}
\textbf{Pass@k}=\underset{\text{Problems}}{\mathbb{E}}\left[1-{\binom{n-c}{k}}/{\binom{n}{k}}\right]
\end{equation}
In Eq.~\ref{eq:passk}, $n$ represents the total number of samples, $c$ denotes the number of correct samples, and $k$ stands for $k$ in $pass@k$. In particular, we calculate both class-level \passk and method-level \passk in class-level code generation tasks: class-level \passk considers code samples at the class granularity and method-level \passk consider code samples at the method granularity. A class-level code sample is deemed correct if it passes all the method-level and class-level test cases; and a method-level sample is deemed correct if it passes all the method-level test cases. In order to maintain an acceptable cost and response time in practical settings, we set $n$ to five. To address the challenge of high sampling variance, we employ an unbiased estimator in line with previous work~\cite{chen2021huamneval}.  

In addition to code correctness, we further measure the model capability of generating code that is dependent to the contexts (\ie{} invoking the other methods declared in the class or assessing the fields in the class). Such capability is essential in class-level code generation. To this end, we design the metric \dep, which calculates the percentage of dependencies generated per method compared to the actual number of dependencies in the canonical solution method. In particular, we consider method dependencies DEP(M) and field dependencies DEP(F): 

\noindent
\begin{minipage}{.5\linewidth}
\begin{equation}
\label{eq:depM}
\footnotesize
% \vspace{-2mm}
\operatorname{\textbf{DEP}}(M)=\frac {\sum_{i=1}^{n}{G_{i}(M)}} {\sum_{i=1}^{n}{S_{i}(M)}}
\end{equation}
\end{minipage}%
\begin{minipage}{.5\linewidth}
\begin{equation}
\label{eq:depF}
\footnotesize
% \vspace{-2mm}
\operatorname{\textbf{DEP}}(F)= \frac{\sum_{i=1}^{n}{G_{i}(F)}} {\sum_{i=1}^{n}{S_{i}(F)}}
\end{equation}
\end{minipage}
$G_{i}(M/F)$ is the number of generated method/field dependencies in the $i^{th}$ method, and $S_{i}(M/F)$ is the number of actual method/field dependencies in the $i^{th}$ method of the canonical solution.

For each generation strategy, we employ nucleus sampling to generate 5 samples and calculate \passk{} metrics with $k=\{{1,3,5}\}$. In addition, we also use the greedy sampling strategy to generate one single greedy sample and calculate \passone and \dep metrics. More sampling details are in Section~\ref{sec:setting}. 

%% file: sections/evaluation/setting.tex
\subsection{Implementation Details}
\label{sec:setting}
We use the OpenAI API interface, specifically the ``gpt-4'' and ``gpt-3.5-turbo'' model interface~\cite{api}, in July 2023. For open-source LLMs, we directly obtain and run their released versions from their official repositories based on the documentation. The maximum window length is set to 2,048 tokens for all LLMs, determined by the smallest maximum window length among the studied LLMs. 

In line with recent work~\cite{yu2023codereval}, we consider two sampling methods for code generation: (i) nucleus sampling~\cite{DBLP:conf/iclr/nsample}, where five solution code samples are randomly generated for each task with a temperature of 0.2~\cite{chen2021huamneval} and default top\_p, and (ii) greedy sampling~\cite{DBLP:journals/tsp/gsample}, where only one single solution code sample is generated for each task using greedy decoding, \ie setting the ``do\_sample'' hyperparameter to false (temperature of 0). During each iteration in incremental and compositional generation, we obtain the Top-1 generated result for each method.
Our experiments are run on a computational infrastructure comprising eight A800-80G GPUs.

%% file: sections/results.tex
\input{sections/rq1}

\input{sections/rq2}
\input{sections/rq3}

\input{sections/rq4}

\input{sections/rq5}

%% file: sections/rq1.tex
\input{tables/nucleus_sampling}
\section{Results}
\subsection{RQ1: Overall Correctness}\label{sec:eval:rq1}
Figure~\ref{fig:pass1bar} shows the class-level and method-level \passone{} with greedy sampling of studied LLMs on \ourbenchmark{} and \humaneval{}. Due to space limits, we only present the best class-level \passone{} (and corresponding method-level \passone{}) for each model among the three generation strategies. A detailed comparison among three  generation strategies is discussed in Section~\ref{sec:eval:rq3}. Method-level \passone{} results on \humaneval{} are directly adopted from the latest work~\cite{luo2023wizardcoder}, and \chatglm{} results on  \humaneval{} are absent from  existing evaluation. Table~\ref{table:nucleus_sampling_results} presents the class-level and method-level \passk{} with nucleus sampling on \ourbenchmark{}. Similarly, due to space limits, we only present results for the generation strategy with the highest class-level \passone{}. Based on Figure~\ref{fig:pass1bar} and Table~\ref{table:nucleus_sampling_results}, we have the following observations.

\input{image/rq1_pass1}
\parabf{Class-level code generation v.s. Method-level code generation.}
Based on Figure~\ref{fig:pass1bar}, we observe a significant decrease in correctness for all studied models on our class-level benchmark \ourbenchmark{} compared to the existing method-level benchmark \humaneval{}. 
In particular, the best-performing models GPT-4 and GPT-3.5 achieve 85.4\%/68.9\% correctness on method-level tasks in \humaneval{}, but only 37.0\%/27.0\% correctness on class-level tasks in \ourbenchmark{}. Similar trends can be observed on other models, \eg{} \wizardcoder{} correctly generates 59.8\% methods on \humaneval{}, but only 11.0\% correct classes in our benchmark. 
Despite the inherent challenges of generating a class with multiple methods, the observed decrease in correctness on our benchmark \ourbenchmark{} is not solely due to the larger number of methods to generate. The code generated by all models also shows lower method-level correctness on \ourbenchmark{} compared to \humaneval{}.
For instance, the method-level \passone{} of \gptfour{} and \gptthree{} drops from  85.4\%/68.9\% (on \humaneval) to 62.5\%/52.5\% (on \ourbenchmark{}). This drop could be attributed to the complexity of generating code that depends on other context, which is known to be more challenging than generating standalone code. This finding is consistent with recent work~\cite{yu2023codereval}. In summary, our results show that existing LLMs still have limited performance in solving complicated coding tasks, such as class-level code generation. 

In addition, we observe that the model performance in the standalone method-level code generation tasks does not necessarily reflect their capability of class-level code generation. For example, while \wizardcoder{} and \starcoder{} exhibit much higher method-level \passone{} (59.8.4\% and 34.1\%) compared to \santacoder{} (14.6\%) on \humaneval, all three model exhibit similar performance on class-level code generation tasks in \ourbenchmark{} (around 10\% - 11\% \passone{}). This indicates that the method-level coding ability cannot equivalently represent the class-level coding ability among LLMs, further confirming the necessity of building a class-level code generation benchmark. 

\finding{Existing LLMs demonstrate substantially lower performance on class-level code generation tasks compared to standalone method-level code generation tasks. Additionally, the method-level coding ability cannot equivalently represent the class-level coding ability among LLMs. These findings strongly confirm the motivation and necessity of constructing class-level code generation benchmarks.}

\parabf{Comparison among models.} 
As shown in Figure~\ref{fig:pass1bar} and Table~\ref{table:nucleus_sampling_results}, GPT series (\gptfour{} and \gptthree{}) substantially outperform all the other models on solving class-level coding tasks with both greedy sampling and nucleus sampling. For example, in Table~\ref{table:nucleus_sampling_results}, they outperform the third-ranked model \wizardcoder{} by 25.4\% and 17.4\% in class-level \passone{} with nucleus sampling. Such results indicate the relatively stable dominance of GPT models when generalized to solve more challenging class-level coding tasks. 

The second-ranked tier includes larger code models like \starcoder{}, \inscodegen{}, and \wizardcoder{}, achieving similar \passone{} with greedy sampling ranging from 10.0\% - 11.1\%. Notably, while these models show significant performance differences on method-level coding tasks in \humaneval{} (\wizardcoder{} outperforms \inscodegen{} by 27.5\% on \humaneval{}), they perform similarly on class-level coding tasks. Smaller models (\eg{} \polycoder{}) or general models (\eg{} \chatglm{}) often exhibit worse performance, as expected due to the importance of model size and instruction datasets for generalization. The only exception is \santacoder{}, which achieves comparable performance to larger code models (\starcoder{}, \wizardcoder{}, and \inscodegen{}) with a much smaller model size.

\finding{On class-level code generation, \gptfour{}/\gptthree{} still exhibits dominate superior than other LLMs; \starcoder{}, \inscodegen{}, and \wizardcoder{} perform similarly as the second tier; small or general models often perform the worse, except \santacoder{}, which achieves comparable performance to larger models but with much less parameters.}

%% file: tables/nucleus_sampling.tex
\begin{table}[htp]
    % \footnotesize
    \centering
    \caption{\passk with Nucleus Sampling on \ourbenchmark }
    \vspace{-2mm}
    \label{table:nucleus_sampling_results}
    % \vspace{-4mm}
    \begin{adjustbox}{width=1\columnwidth}
    \begin{tabular}{c|ccc|ccc}
        \hline
        \multirow{2}{*}{\textbf{Model}} & 
        \multicolumn{3}{c|}{\textbf{Class-level}} & 
        \multicolumn{3}{c}{\textbf{Method-level}} 
         \\ \cline{2-4} \cline{5-7}
         & Pass@1 & Pass@3 & Pass@5 & Pass@1 & Pass@3 & Pass@5\\ \hline
        GPT-4 &\textbf{37.6\%}& \textbf{41.3\%}&\textbf{42.0\%}&\textbf{62.8\%}& \textbf{67.4\%}&\textbf{68.5}\%\\ \hline
        GPT-3.5 &29.6\%& 34.9\%&36.0\%&50.4\%& 59.0\%&61.1\%\\ \hline
        WizardCoder&12.2\%& 20.0\%&23.0\%&35.2\%&47.1\%&51.1\%\\ \hline  
        Instruct-StarCoder  & 10.2\% & 12.7\% & 14.0\%  & 23.1\% & 26.5\% & 27.7\% \\ \hline
        SantaCoder &8.6\%&9.9\% &10.0\% &27.7\%&33.0\%&34.9\%\\ \hline
        Instruct-CodeGen &8.2\%&12.3\%&13.0\%&24.9\%&34.3\%&37.1\%\\ \hline
        CodeGeeX &7.2\%&9.4\% &10.0\% & 21.2\%&27.1\%&29.5\% \\ \hline
        InCoder &6.2\%&7.6\% & 8.0\%&21.1\%&26.5\%& 29.1\% \\ \hline   
        Vicuna &3.0\%&3.6\% &4.0\%&11.0\%&15.8\%&18.4\% \\ \hline
        ChatGLM&1.4\%&2.6\%& 3.0\%&8.2\%&11.2\%&12.4\% \\ \hline
        PolyCoder &1.4\%&2.2\% &3.0\%&13.2\%&17.5\%&19.6\% \\ \hline
        
    \end{tabular}
    \end{adjustbox}
\end{table}

%% file: image/rq1_pass1.tex
\begin{figure}[htb]
	\centering
        % \vspace{-2mm}
	\includegraphics[width=0.75\columnwidth]{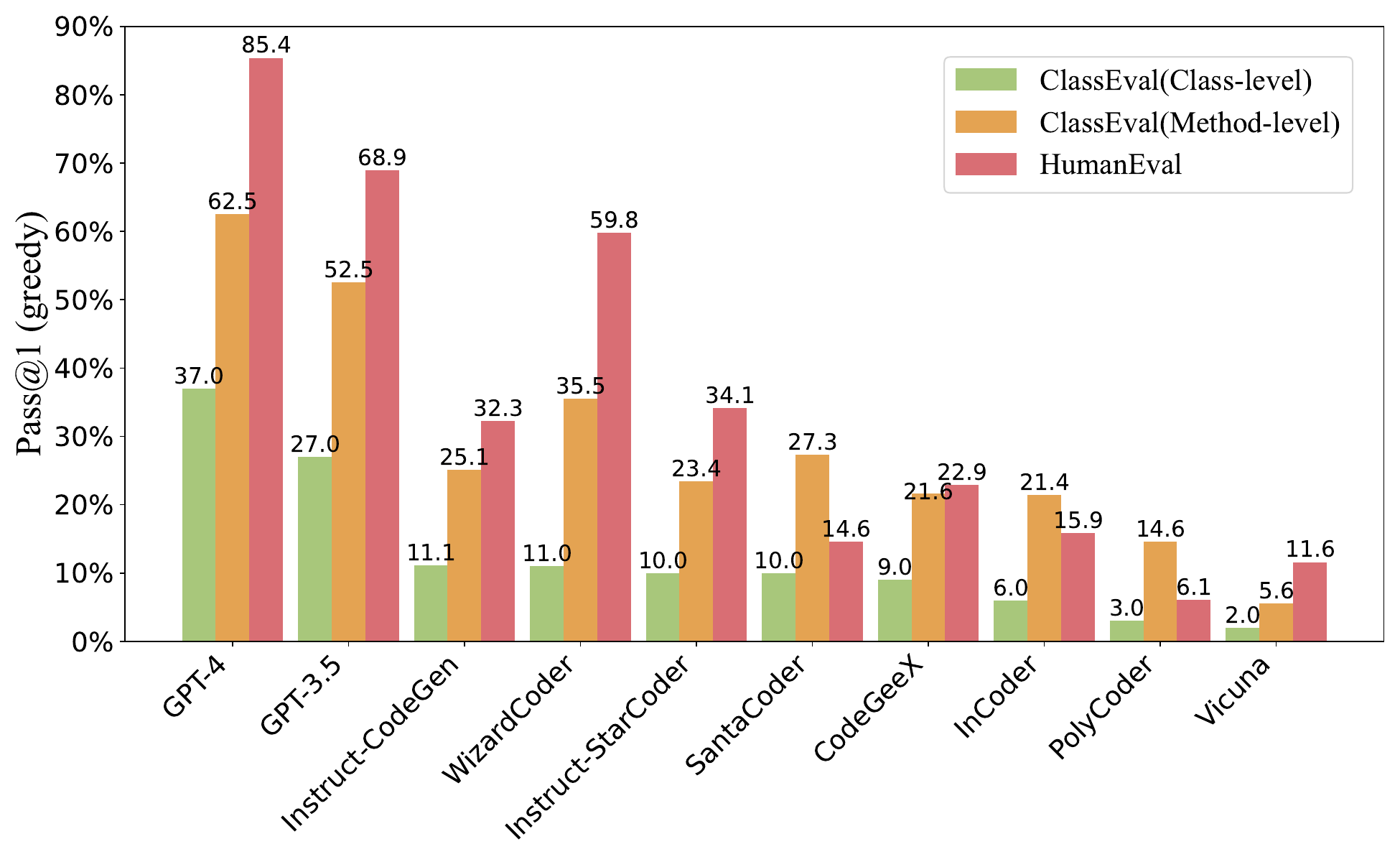}
        \vspace{-2mm}
	\caption{\passone (greedy) on \ourbenchmark{} and \humaneval{}} 
	\label{fig:pass1bar}
        
\end{figure}

%% file: sections/rq3.tex
\input{image/generation_strategies}

\subsection{RQ2: Generation Strategies}\label{sec:eval:rq2}
Figure~\ref{fig:generationapproach} compares the class-level \passfive{} and method-level \passfive of three different generation strategies (\ie{} holistic, incremental, and compositional generation). Based on the figure, overall, we find that the best generation strategy varies among different LLMs.

\parabf{Holistic strategy v.s. others.} On one hand, holistic generation is the best generation strategy only for the two models GPT-4 and GPT-3.5, which achieves much higher class-level \passfive than the other two strategies (\ie{} the improvements range from 6\% to 9\% for GPT-4 and 4\% to 14\% for GPT-3.5). In addition, even for the method-level correctness, holistic generation still outperforms generating method in an incremental or compositional way (\ie{} 1.4\% - 9.0\% improvement in method-level \passfive{}). On the other hand, the trends are different for the other models, which actually perform much better when generating the class method by method, namely with the incremental or compositional  strategies. For example, in terms of the class-level correctness, \codegeex{} and \santacoder{} generate 9\% and 7\% more correct classes with the incremental strategy compared to the holistic generation strategy. The main reason is that these models are able to generate much more correct methods (\ie{} 27.9\% and 19.2\% higher method-level \passfive{}) when generating each method in separate iterations
compared to generating all methods at once. Therefore, these models have higher chance to generate more correct classes if they are able to generate more correct methods with the incremental or compositional strategy.

One potential reason for the observation above might be that most models (except GPT ones), exhibit rather limited capability of utilizing long input contexts, thus finding it more challenging to fully understand the code generation tasks given the entire class skeleton. As revealed by the recent work~\cite{DBLP:journals/corr/abs-2307-03172}, LLMs often become substantially less effective with the increasing length of inputs; and in particular they tend to make better usage of the information located at the beginning or end of the inputs than that in the middle of inputs. Therefore, most existing LLMs perform better in generating a class method by method, since the task inputs are with the more atomic focus in such an incremental or compositional generation scenario; for models like \gptthree{} and \gptfour{} with a better understanding of long instructions, feeding the class-level context all at once is actually beneficial for them to fully capture and utilize the constraints between each method, leading to better class-level code correctness.

\parabf{Incremental strategy v.s. compositional strategy.} 
As for the two method-by-method strategies (\ie{} incremental and compositional strategies), we find the studied models actually have different preference on them. In particular, compared to the compositional generation manner, the additional inputs (the method body generated in previous iterations) in the incremental strategy are helpful for some models such as \inscodegen{}, \incoder{}, \codegeex{}, and \santacoder{}. In contrast, the previously-generated method bodies can negatively affect the performance of models like \starcoder{} and \wizardcoder{}, resulting in a lower class-level correctness in  incremental generation. In addition to the limited capability of handling long inputs mentioned above, another potential reason for the model's preference on a rather individual generation manner might be that the compositional generation aligns better with simple and atomic task instructions during instruction tuning. 

\finding{Generating the entire class all at once (\ie{} holistic  strategy) is the best generation strategy only for \gptfour{} and \gptthree{}. For the other models, method-by-method generation (\ie{} incremental and compositional) works better. Such a disparity might stem from their limited capability of understanding the long instructions and utilizing the middle information. }

%% file: image/generation_strategies.tex
\begin{figure}[htbp]
    \centering 
    \vspace{-3mm}
    \subfigure[Class-level \passfive{}]{\label{fig:pass5subfig1}
    \includegraphics[width=0.23\textwidth]{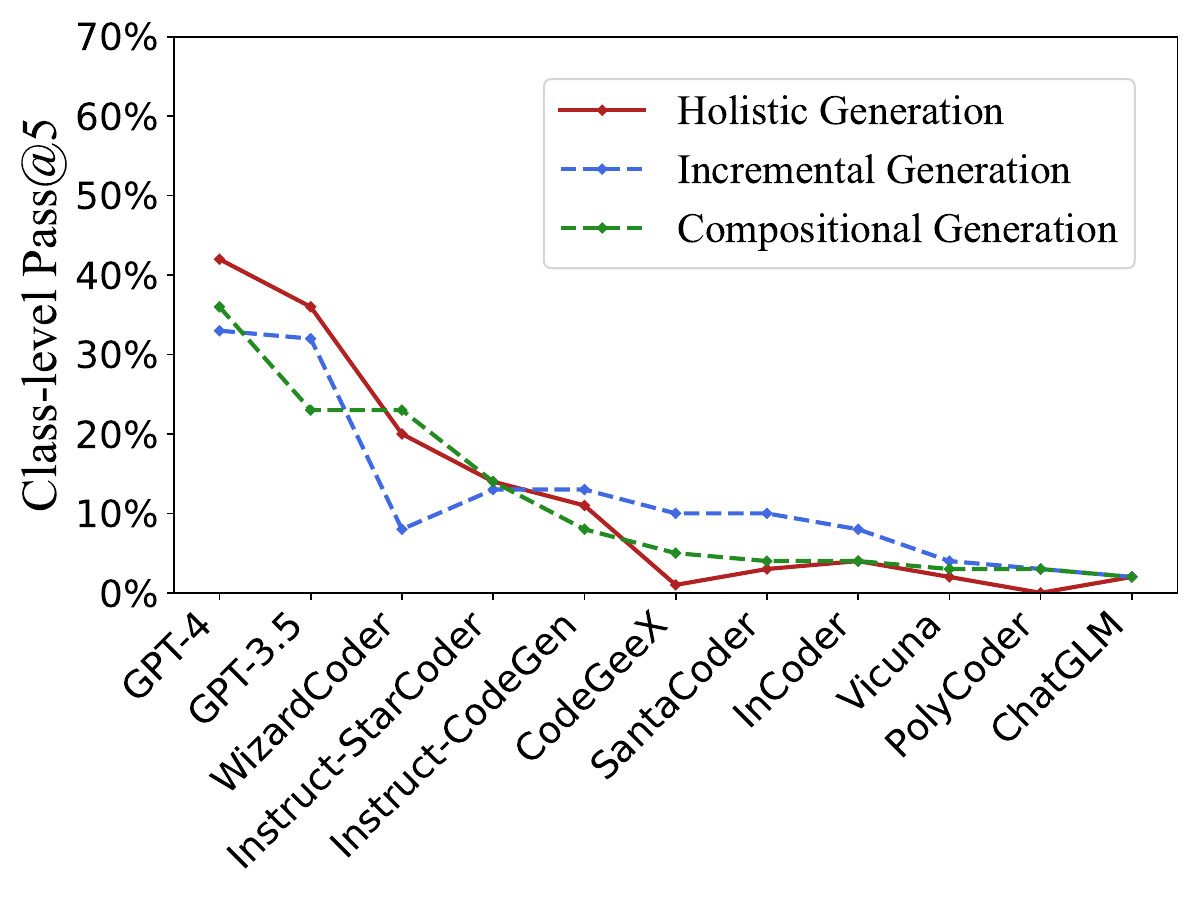}}
    % \hspace{2cm} 
    \vspace{-1mm}
    \subfigure[Method-level \passfive]{\label{fig:pass5subfig2}
    \includegraphics[width=0.23\textwidth]{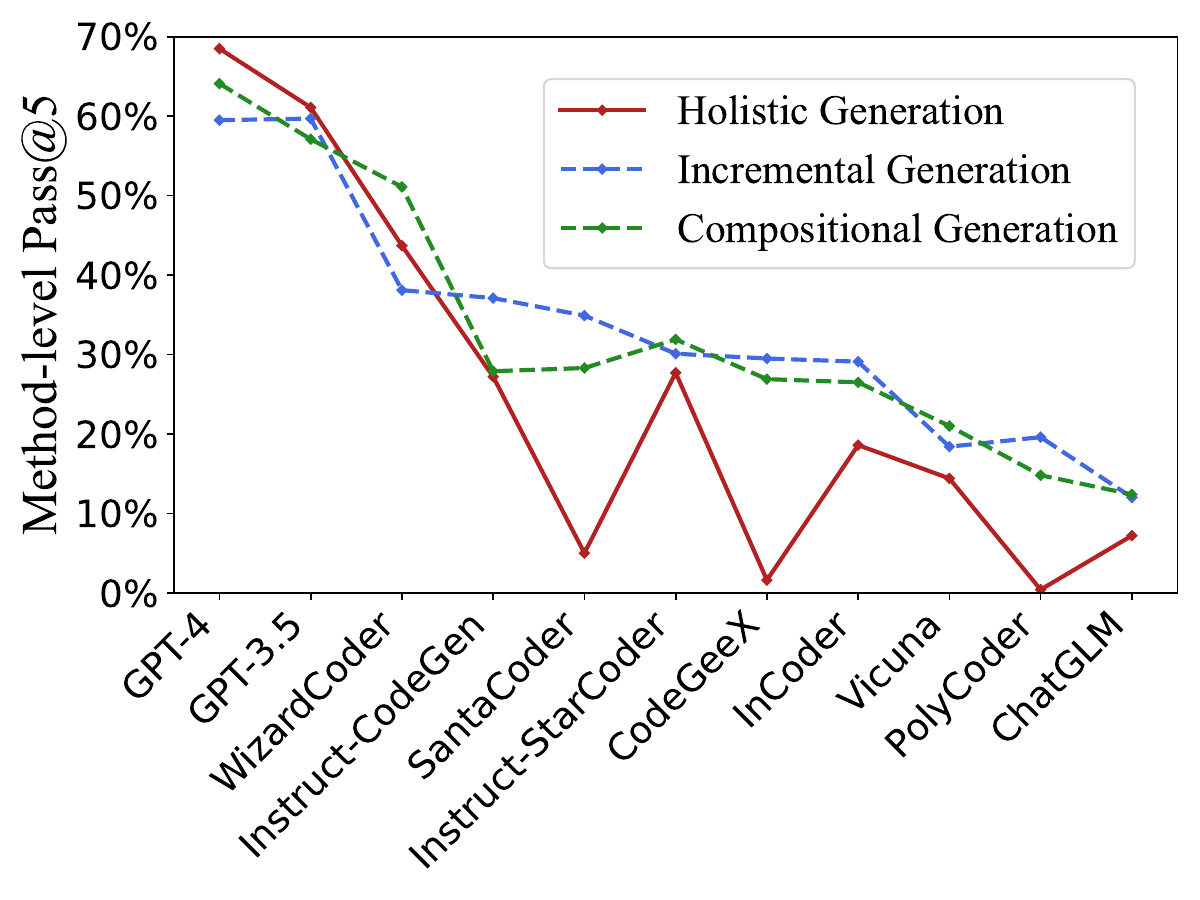}}
    \vspace{-2mm}
    \caption{\passfive of Three Generation Strategies}
    \label{fig:generationapproach}
\end{figure}

%% file: sections/rq4.tex
\subsection{RQ3: Dependency Generation}\label{sec:eval:rq3}
\parabf{Method dependency v.s. Field dependency.} Figure~\ref{fig:dep} presents the average field dependencies DEP(F) and the method dependencies DEP(M) of each model with the nucleus sampling. For space limits, we only present the best results among three generation strategies. Based on Figure~\ref{fig:dep}, we can find that all models exhibit a much higher success rate in generating code dependent to fields than generating code dependent to other methods (\ie{} higher DEP(F) than DEP(M) on all the models). In other words, it might be much easier for models to generate field-accessing code than method-invoking code. In addition, among all the models,  GPT models still show consistent superior in generating dependent code, \eg{} GPT-4 substantially outperform other LLMs by at least 12.6\%/6.3\% improvement in DEP(F)/DEP(M).  

\input{image/dep}

Given our observation above that it is more challenging to generate method dependency, we further investigate how each model performs at correctly generating code that invokes different number of other methods. Figure \ref{fig:dep_stack} is a stacked-bar plot that show the ratio of correctly-generated methods to all methods with the given number (\ie{} 0, 1, 2) of method dependencies (based on the canonical solution). Based on the figure, we can find that all the models perform best when generating methods that do not invoke any other method declared in the class (the blue bar in the figure). In addition, we find that no obvious difference when most models generate code invoking one other method (the green bar) or invoking two other methods (the yellow bar). In particular, for all the models, the average ratio of correctly-generated code that invokes one or two method(s) is 27.7\% and 27.6\% respectively. 

\input{image/dependency_stacked}

\finding{It is easier for all the models to generate field-accessing code than method-invoking code. Additionally, they  are better at generating standalone methods that do no invoke any other method.}

%% file: image/dep.tex
\begin{figure}[htb]
	\centering
        %\vspace{-2mm}
	\includegraphics[width=0.65\columnwidth]{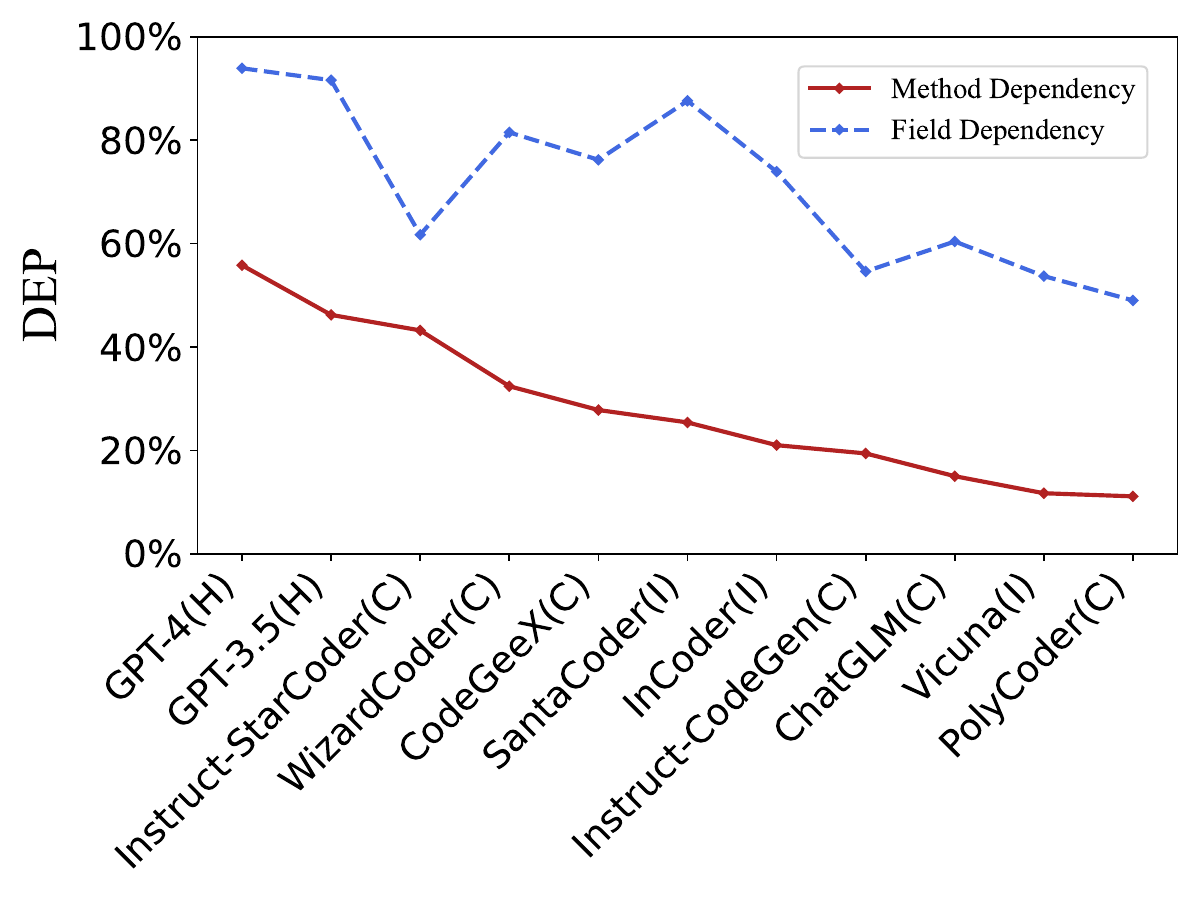}
        \vspace{-2mm}
	\caption{DEP(F) and DEP(M) in Nucleus Sampling} 
	\label{fig:dep}
        % \vspace{-4mm}
\end{figure}

%% file: image/dependency_stacked.tex
\begin{figure}[htb]
	\centering
        %\vspace{-2mm}
	\includegraphics[width=0.70\columnwidth]{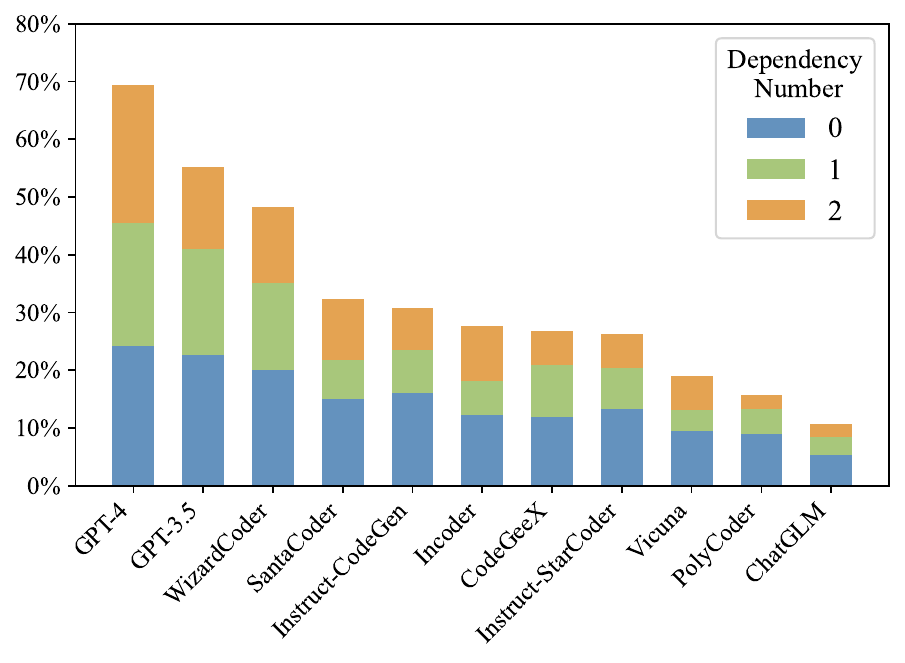}
        \vspace{-2mm}
	\caption{Distribution of correctly-generated methods in increasing method dependencies}
	\label{fig:dep_stack}
        %\vspace{-4mm}
\end{figure}

%% file: sections/rq5.tex
\subsection{RQ4: Bad Case Analysis}\label{sec:eval:rq4}
We further analyze the incorrectly-generated classes. To this end, we automatically parse the error logs generated during interpretation and execution, and present the error distribution of all models in Figure~\ref{fig:error}. In particular, we find that most incorrect code encounters \textit{AttributeError} and \textit{TypeError}, indicating the limited model ability of understanding and satisfying syntactic or semantic constraints in the code context. Additionally, a few cases encounter \textit{KeyError} due to erroneous operations on the dictionary variable. Figure~\ref{fig:error_code} shows such an example from GPT-3.5, resulting from a misinterpretation of the field dependency. Specifically, the model erroneously accesses the first element of the field \textit{BMI\_std} list, which is a dictionary with the key ``male''. Attempting to access the key \textit{self.sex} as ``female'' within this dictionary triggers a KeyError. This case indicate one of the challenges that LLMs might encounter in handling inherent class-level dependencies.

\input{image/error_example}

\finding{The classes generated by LLMs suffer from AttributeError and TypeError most frequently. In addition, the models might encounter difficulties in understanding the dependent contexts in the class.}

%% file: image/error_example.tex
% \begin{figure}[htb]
% 	\centering
%         %\vspace{-2mm}
% 	\includegraphics[width=0.7\columnwidth]{images/error example.pdf}
%         % \vspace{-2mm}
% 	\caption{Incorrectly-Generated Code Example of KeyError} 
% 	\label{fig:error_code}
%         % \vspace{-4mm}
% \end{figure}

\begin{figure}
  \centering
  \begin{minipage}[t]{0.5\columnwidth}
    \centering
    \includegraphics[width=\textwidth]{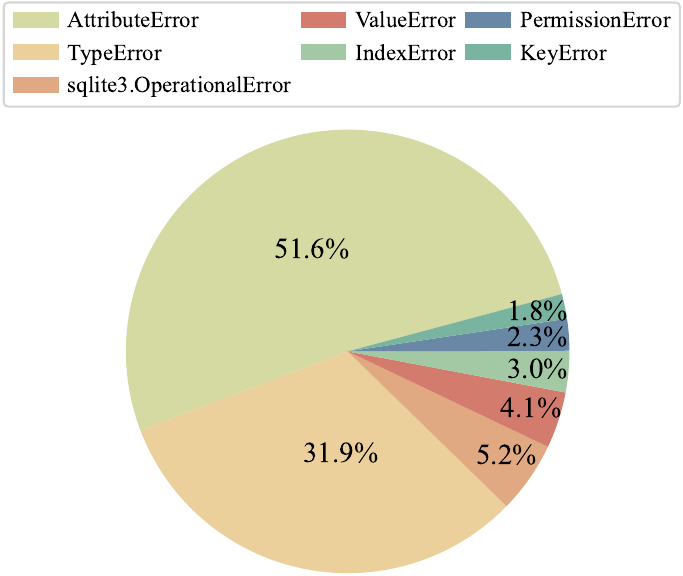}
    \captionsetup{font=footnotesize}
    \caption{Error Distribution }
    % \todo{update the fig}}
    \label{fig:error}
  \end{minipage}
  \begin{minipage}[t]{0.45\columnwidth}
    \centering
    \includegraphics[width=\textwidth]{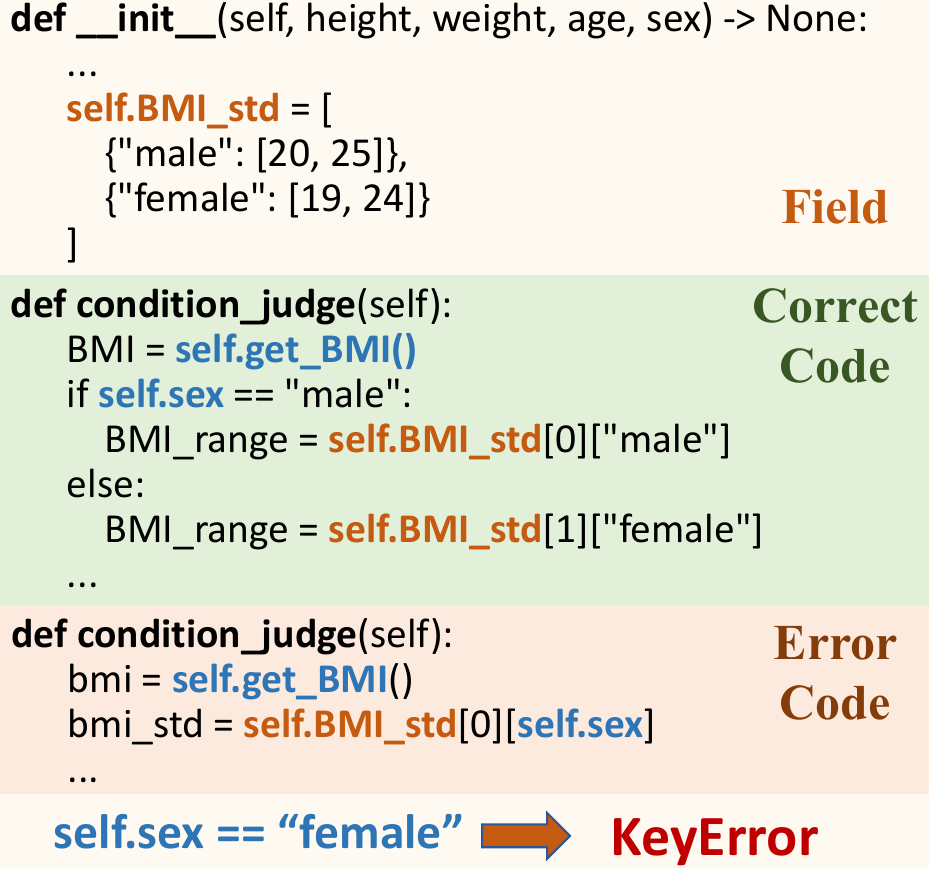}
    \captionsetup{font=footnotesize}
    \caption{KeyError Code Example }
    \label{fig:error_code}
  \end{minipage}
\end{figure}

%% file: sections/threat.tex
\section{Threats to Validity}
\textbf{Threats in benchmark construction.} 
One potential threat is the data leakage between our benchmark and model training data, thus we \textit{manually} construct the benchmark \ourbenchmark{}. We also involve multiple  participants to mitigate the subjectiveness and mistakes in manual participation. Another threat lies in the limited size and programming languages in our current benchmark, which cannot guarantee the generalizability of our findings, and we plan to continually extend our benchmark in the future. 
\parabf{Threats in empirical study.}
To avoid buggy model implementation, we adopt the public versions following official guidelines of each model. Another threat lies in the prompts used in our experiments, which might impact our findings. To avoid underestimating  studied models, we perform a pilot study on a small set of prompt candidates and select the one with the best performance on three separate class-level coding tasks. 
We also report the results with greedy decoding, which is deterministic, so as to mitigate the randomness in  model responses.

%% file: sections/related.tex
\section{Related Work}

Since we have discussed most relevant work on  LLMs and existing code generation benchmarks in Section~\ref{sec:back}, we mainly introduce related work on LLM evaluation in this section. 
Multi-faceted evaluation for LLMs is crucial for understanding the model capabilities given the black-box nature of LLMs. To date, the evaluation for LLMs has covered a wide range~\cite{chang2023survey}, encompassing not only traditional NLP tasks (\eg{} sentiment analysis~\cite{bang2023multitask}, question answering~\cite{bai2023benchmarking}, and reasoning~\cite{bian2023chatgpt}) but also some specific downstream domains (\eg{} medicine~\cite{chervenak2023promise}, agent~\cite{huang2023language}, and recommendation system~\cite{fan2023recommender}). Specifically in software engineering domain, current evaluation focuses primarily on code generation tasks~\cite{chen2021huamneval,austin2021mbpp,li2023enabling,liu2023humanevalplus}. 
Many code LLMs (\eg{} Codex~\cite{chen2021huamneval} and PanGu-Coder2~\cite{shen2023pangucoder2}) are released along with its rigorous evaluation on \humaneval{} to demonstrate their capabilities on code generation. 
While these previous efforts do not take scenarios beyond function-level code generation into account, our work fills this gap by manually constructing the first class-level code generation benchmark for evaluating LLM on  more complicated and practical software development tasks. 

%% file: sections/conclusion.tex
\section{Conclusion}
This work makes the first attempt to evaluate LLMs on class-level code generation. We first manually construct the first class-level code generation benchmark \ourbenchmark{} and perform the first study of 11 state-of-the-art LLMs on class-level code generation. We find that all LLMs perform much worse on class-level code generation compared to the method-level. While GPT models still dominate other LLMs on class-level code generation, the ranking of model performance on method-level code generation no longer holds in the class-level code generation. Besides, most models (except GPT models) perform better when generating the class method by method; and they have the limited ability of generating dependent code.